%% file: main.tex
\documentclass[conference]{IEEEtran}
\IEEEoverridecommandlockouts

\input{macros}

\begin{document}

\title{Adaptive Depth Aggregation in Graph Transformers for Long-Range Graph Reasoning}

\author{
\IEEEauthorblockN{Yifeng Lu}
\IEEEauthorblockA{School of Computer Science and Engineering\\
Sun Yat-sen University\\
Guangzhou, China\\
luyf55@mail2.sysu.edu.cn}
\and
\IEEEauthorblockN{Zhen Wang}
\IEEEauthorblockA{School of Computer Science and Engineering\\
Sun Yat-sen University\\
Guangzhou, China\\
wangzh665@mail.sysu.edu.cn}
}

\maketitle

\begin{abstract}
Graph Transformers improve long-range node-to-node communication through
global or structure-aware interactions.
However, improving cross-node communication does not determine 
how computations produced at different depths should be combined.
Standard residual streams accumulate successive computations with fixed unit coefficients shared across
nodes. This fixed aggregation is a residual-interface constraint: once
intermediate contributions have been produced, the interface cannot explicitly
reweight them by node. We hypothesize that this constraint can create a
\emph{depth-wise aggregation mismatch} when the relative utility of these
intermediate contributions varies across nodes and structural contexts. 
To study this hypothesis, we formalize Graph Transformer residual computation
as a causal node-wise history of graph-propagation and feed-forward network (FFN) contributions,
making depth aggregation an explicit interface choice. We realize
node-adaptive aggregation over this history with Attention Residuals (AttnRes)
while preserving the backbone's graph-propagation and auxiliary-state machinery.
Block AttnRes reduces the routing cost of long histories relative to Full
AttnRes by pooling consecutive contributions. We instantiate the interface in GPS and GRIT and evaluate it on
all five ECHO tasks. Both variants achieve lower mean absolute error (MAE) than their
independently tuned backbones on every task, and at least one obtains the
lowest mean among all compared methods in each task. Routing-only donor-swap
interventions support the mismatch hypothesis, while additional ablations
examine the key design choices. Together, these results support node-adaptive
depth aggregation as a useful complement to cross-node communication in
long-range graph reasoning.
\end{abstract}

\begin{IEEEkeywords}
adaptive depth aggregation, Attention Residuals, Graph Transformers,
long-range graph learning, residual connections.
\end{IEEEkeywords}

\input{sections/01_introduction}
\input{sections/02_related_work}
\input{sections/03_method}
\input{sections/04_experimental_setup}
\input{sections/05_main_results}
\input{sections/06_ablation_analysis}
\input{sections/07_conclusion}

\IEEEtriggeratref{35}
\bibliographystyle{IEEEtran}
\bibliography{refs}

\end{document}

%% file: macros.tex
\usepackage{cite}
\usepackage{amsmath}
\usepackage{amssymb}
\usepackage{amsfonts}
\usepackage{booktabs}
\usepackage{graphicx}
\usepackage{textcomp}
\usepackage{xcolor}
\usepackage[hidelinks]{hyperref}

\def\BibTeX{{\rm B\kern-.05em{\sc i\kern-.025em b}\kern-.08em
    T\kern-.1667em\lower.7ex\hbox{E}\kern-.125emX}}

\newcommand{\method}{AttnRes-GT}
\newcommand{\gpsattnres}{GPS+AttnRes}
\newcommand{\gritattnres}{GRIT+AttnRes}
\newcommand{\echosynth}{ECHO-Synth}
\newcommand{\echochem}{ECHO-Chem}
\newcommand{\echobenchmark}{ECHO benchmark}

%% file: sections/01_introduction.tex
\section{Introduction}

Many graph-learning problems involve dependencies spanning distant graph
regions, including shortest-path reasoning and molecular prediction settings
designed to test long-range propagation~\cite{NEURIPS2022_8c3c6668,echobenchmark}.
Message-passing graph neural networks (GNNs) construct such interactions through
repeated local aggregation~\cite{pmlr-v70-gilmer17a}. 
Distant signals must consequently traverse long computational paths and may encounter structural
bottlenecks~\cite{alon2021bottleneck,topping2022oversquashing}.
Graph Transformers improve this \emph{node-to-node communication} with global or
structure-aware node interactions, allowing nonlocal information to be
acquired in fewer propagation steps~\cite{NEURIPS2021_b4fd1d2c,NEURIPS2022_5d4834a1,pmlr-v202-ma23c}.
We focus on Graph Transformer architectures in which graph-propagation 
and node-wise feature-transformation sublayers are organized through standard residual connections.

Improving cross-node communication does not by itself determine how
computations produced at different stages should be reused by later stages.
Nonlocal information acquired at one stage may subsequently be transformed,
integrated, or refined by graph-propagation and feed-forward network (FFN) sublayers. These
transformations unfold across network depth, while the residual interface
determines how their contributions are combined.

Standard residual accumulation~\cite{He_2016_CVPR} uses a fixed rule for this
combination. If $r_t$ is the residual state after sublayer $t$ and $h_t$ its
new contribution, then
$r_t=r_{t-1}+h_t=r_0+\sum_{i=1}^{t}h_i$.
The contributions $h_i(v)$ remain node- and context-dependent, but their
aggregation coefficients are fixed to one for every node. This cumulative
path supports optimization, and the backbone can adapt what each stage writes
into $h_i(v)$. The constraint considered here is narrower: once these
contributions have been produced, standard residual accumulation exposes them
to subsequent computation through the same fixed unit coefficients.
We hypothesize that this constraint can create a \emph{depth-wise aggregation
mismatch} when nodes benefit from different mixtures of earlier contributions.
Although this limitation can arise more broadly, 
it may become especially consequential in long-range tasks 
when a prediction depends on coordinating information formed across multiple stages of computation.
This does not imply that distant nodes universally prefer earlier or later contributions; the useful mixture can
itself be node- and context-dependent.

To study this hypothesis, we organize the embedded input and successive
graph-propagation and FFN contributions as a causal history for each node. A
sublayer reads the available contributions and appends its new contribution
after the transformation. This exposes depth aggregation as an explicit
interface choice and motivates a node-conditioned alternative.

\begin{figure*}[!t]
  \centering
  \includegraphics[width=\textwidth,trim=4bp 1bp 72bp 1bp,clip]{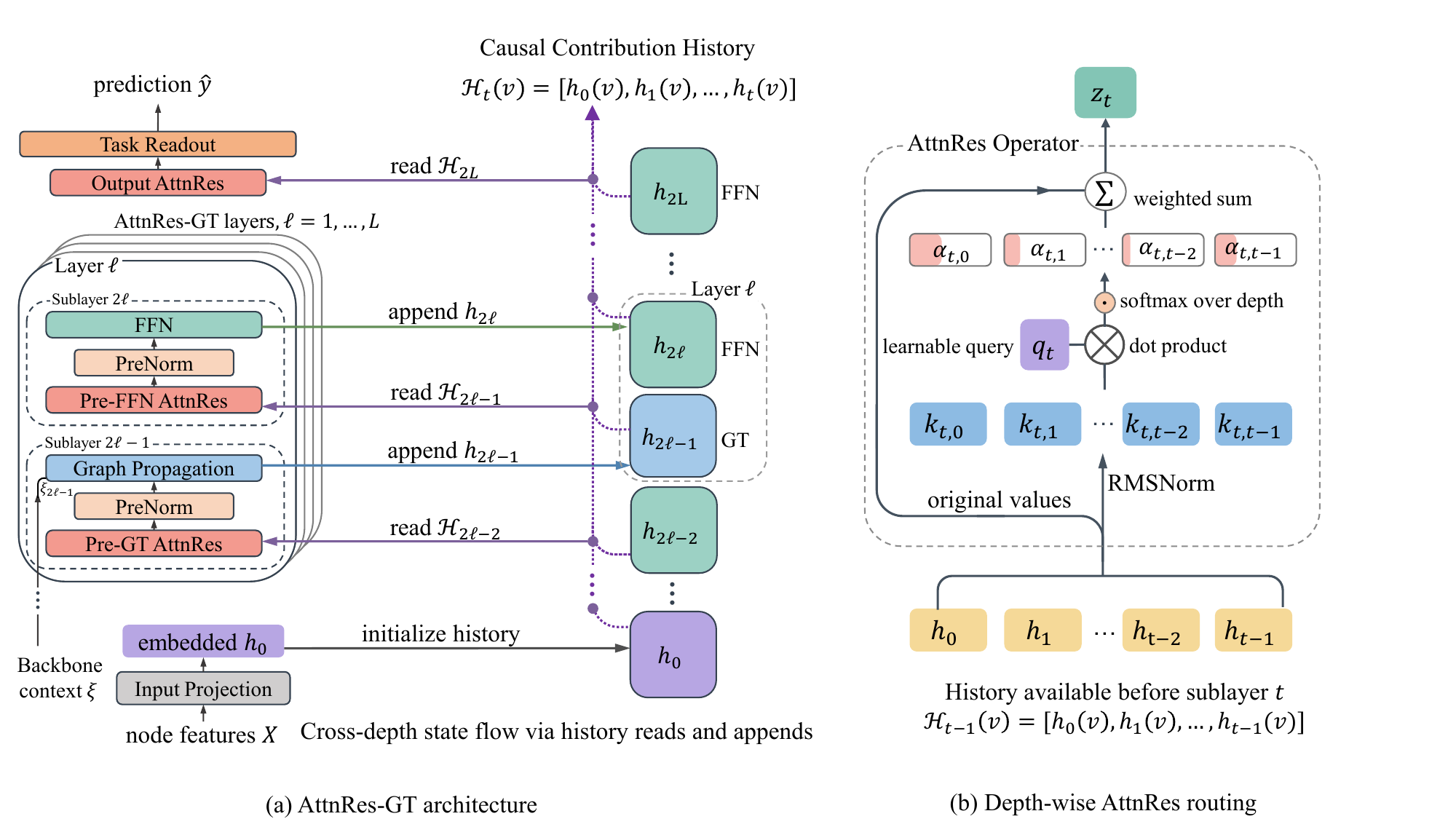}
  \caption{Adaptive depth aggregation in \method{}. (a) Each layer performs Pre-GT and Pre-FFN
  causal history reads before appending the new graph-propagation and FFN
  contributions; Output AttnRes routes the complete history to the unchanged
  readout. (b) The AttnRes realization uses RMS-normalized history candidates as keys and the original
  candidates as values. A module-specific query shared across nodes produces
  node-specific depth weights. Full AttnRes is shown; Block AttnRes sum-pools
  consecutive contributions as defined in Section~\ref{sec:block-attnres}.}
  \label{fig:attnres-gt-architecture}
\end{figure*}

Attention Residuals (AttnRes) provide a natural routing operator for this
purpose through content-dependent attention over preceding sublayer
outputs~\cite{chen2026attnres}. We use AttnRes to instantiate the adaptive
interface and refer to the resulting design as \method{}.
We then ask whether node-dependent depth routing can reduce the
hypothesized mismatch in Graph Transformers with standard residual accumulation.
The backbone graph-propagation mechanism governs cross-node communication, 
while the \method{} interface governs how each node reuses computations across depth.
Figure~\ref{fig:attnres-gt-architecture} illustrates the read--transform--append recurrence.
We instantiate it in GPS, which combines local message passing with global attention, and GRIT, which
uses structure-aware attention, yielding \gpsattnres{} and
\gritattnres{}~\cite{NEURIPS2022_5d4834a1,pmlr-v202-ma23c}.

We evaluate both instantiations on the full \echobenchmark{}, spanning
synthetic long-range reasoning in \echosynth{} and molecular prediction in
\echochem{}~\cite{echobenchmark}. Under matched search and evaluation
protocols, both variants achieve lower mean MAE than their corresponding
independently tuned backbones on all five tasks. Compared with all results
reported by ECHO, at
least one AttnRes variant attains the lowest mean in each task.
Controlled routing interventions further support this interpretation: learned
depth mixtures are functionally non-interchangeable across nodes and vary
systematically across target-defined structural regimes. Additional studies
examine history fusion, placement, distance-stratified performance, and
computational cost.

Our contributions are:
\begin{itemize}
  \item We analyze a limitation of standard residual accumulation in Graph
  Transformers: intermediate contributions are combined across depth with
  fixed unit coefficients shared across nodes. We hypothesize that this rule
  can create a depth-wise aggregation mismatch when their utility varies
  across nodes and contexts.
  \item We formalize graph-propagation and FFN contributions as a causal
  node-wise history and construct an adaptive depth-aggregation interface that
  retains backbone-specific graph operations and auxiliary-state updates.
  We refer to the resulting design as \method{}, with AttnRes realizing its routing operator.
  \item We evaluate GPS and GRIT instantiations on all five ECHO tasks and use
  controlled routing interventions to test whether learned depth mixtures are
  functionally interchangeable across nodes and structural regimes. Both
  achieve lower mean MAE than their independently tuned backbones on every
  task.
\end{itemize}

%% file: sections/02_related_work.tex
\section{Related Work}

\subsection{Long-Range Graph Learning}

An $L$-layer one-hop message-passing neural network (MPNN) communicates within an $L$-hop neighborhood, where
distant signals may cross structural bottlenecks and compete for fixed-width
representations~\cite{pmlr-v70-gilmer17a,alon2021bottleneck,
topping2022oversquashing}. These observations motivate separating two aspects
of deep graph reasoning: communicating information across the graph and
aggregating computations produced across network depth.
The Long Range Graph Benchmark (LRGB) and ECHO evaluate long-range graph
learning across synthetic, visual, and molecular tasks~\cite{NEURIPS2022_8c3c6668,echobenchmark}.

Communication-path approaches shorten or widen message routes through curvature-based or
spectral rewiring and distance-aware DRew~\cite{topping2022oversquashing,
karhadkar2023fosr,pmlr-v202-gutteridge23a}. Multi-hop and diffusion operators
expose multiple propagation radii~\cite{pmlr-v97-abu-el-haija19a,
gasteiger_predict_2019,NEURIPS2019_23c89427}, while global attention enables
nonlocal interactions without traversing every intermediate edge~\cite{NEURIPS2021_b4fd1d2c}.

Deep propagation is also challenged by over-smoothing~\cite{oono2020graph}.
DeepGCNs use residual, dense, and dilated connections~\cite{Li_2019_ICCV},
reversible GNNs reduce activation storage~\cite{pmlr-v139-li21o}, and
dynamical-system models control smoothing or dissipation through oscillator,
antisymmetric, port-Hamiltonian, or coupled antisymmetric dynamics~\cite{pmlr-v162-rusch22a,gravina2023adgn,heilig2025porthamiltonian,
gravina2025swan}.

These methods alter graph paths, propagation, or stability. Our focus is the
depth-wise aggregation rule used after such graph computations have produced
their residual contributions.

\subsection{Graph Transformers}

Graph Transformers use attention to extend communication beyond local message
passing. A central design challenge is to encode graph structure despite the
absence of a canonical sequence order. The original Graph Transformer augments
self-attention with Laplacian positional encodings~\cite{dwivedi2021generalization}.
SAN develops a spectral attention mechanism~\cite{NEURIPS2021_b4fd1d2c}, 
while Graphormer injects shortest-path, edge, and
centrality information directly into attention~\cite{NEURIPS2021_f1c15925}.
GraphiT modulates attention with diffusion kernels~\cite{mialon2021graphit}, 
providing another route for making pairwise
interactions sensitive to graph geometry.

Subsequent architectures also reconsider what attention operates on and how it
is combined with graph computation. SAT represents each node through its rooted
subgraph before attention~\cite{pmlr-v162-chen22r}, whereas TokenGT tokenizes
both nodes and edges and supplies graph-specific identifiers~\cite{NEURIPS2022_5d842367}. 
GPS interleaves local MPNN computation with global
attention~\cite{NEURIPS2022_5d4834a1}; Exphormer uses expander-based sparse
attention to improve scalability~\cite{pmlr-v202-shirzad23a}; and GRIT couples
node states with relative random-walk-aware node-pair representations~\cite{pmlr-v202-ma23c}. 
We use GPS and GRIT because they realize global graph
communication through materially different backbone designs.

Despite their differences, these Graph Transformers primarily redesign
\emph{node-to-node} communication and its structural biases: they
determine which nodes interact and how graph structure governs the
interaction. This can shorten the communication path for nonlocal information,
while leaving a separate design question: how should the resulting
contributions be aggregated across computation depth for each node?
\method{} changes this residual interface without replacing the backbone's graph-propagation mechanism.

\subsection{Cross-Layer Representation Reuse}

Cross-layer connections support optimization and feature reuse. Residual
connections accumulate new contributions with fixed unit
coefficients~\cite{He_2016_CVPR}, dense
connections concatenate earlier features~\cite{Huang_2017_CVPR}, and GCNII
reinjects the initial node representation in deep GCNs ~\cite{pmlr-v119-chen20v}. 
These mechanisms provide cross-layer reuse through routes fixed by the architecture.

Graph models also combine multiple propagation depths. 
Jumping Knowledge Networks aggregate layer outputs~\cite{pmlr-v80-xu18c}; 
DAGNN adaptively combines decoupled propagation stages~\cite{liu2020towards}; and AGDN learns
hop-wise coefficients within graph diffusion~\cite{sun2025agdn}. 
These methods enable adaptive use of multiple propagation depths,
but operate on layer or hop representations at a stack output or within a propagation module.

Recent Transformer variants explore cross-depth information reuse through mechanisms 
such as carrying intermediate attention signals, mixing residual streams, 
and attending to preceding layer representations~\cite{he-etal-2021-realformer,NEURIPS2024_f67449c7,ICLR2025_f1e8ff97,pmlr-v267-heddes25a}. 
AttnRes provides content-dependent attention over residual
history~\cite{chen2026attnres}. Our study uses its Full and Block operators as
the depth-routing realization within a graph-specific residual formulation:
graph-propagation and FFN contributions form a causal same-node history, while
graph structure and backbone-specific auxiliary states continue to be handled
by the original backbone machinery.

%% file: sections/03_method.tex
\section{Adaptive Depth Aggregation with AttnRes-GT}

We formalize Graph Transformer residual computation as a causal contribution
history, identify the fixed-aggregation constraint, and then instantiate an
adaptive interface with AttnRes before graph-propagation and FFN sublayers and at the final output
(Fig.~\ref{fig:attnres-gt-architecture}).

\subsection{Causal Contribution History}

Let $G=(V,E)$ be an input graph with node-feature matrix $X$, and
let $x(v)$ denote the input feature of node $v\in V$. A learned
input projection produces
$
x_0(v)=W_{\mathrm{in}}x(v)+b_{\mathrm{in}}\in\mathbb{R}^d,
$
and we set $h_0(v)=x_0(v)$ as the first history entry.
Backbone-specific structural states remain part of the original graph
propagation and are excluded from the node-wise history (including
GRIT's node-pair states).

For $t\geq1$, $h_t(v)\in\mathbb{R}^d$ denotes the residual
contribution produced by sublayer $t$. We use $\mathcal H_t(v)$
for the ordered history containing $h_0(v)$ and all contributions
produced through sublayer $t$. 
This history provides a contribution-wise representation of the node residual stream:
each subsequent sublayer obtains its input by routing over the causally available history, 
and no separate cumulative node hidden state is maintained.
Throughout this section, $\ell$
indexes Graph Transformer layers, $t$ indexes graph-propagation and
FFN sublayers in causal execution order, and $i,j$
index history candidates or blocks.

An \method{} with $L$ layers contains $2L$ residual sublayers, indexed by
$t\in\{1,\ldots,2L\}$. For $\ell\in\{1,\ldots,L\}$, sublayer $2\ell-1$ is
the graph-propagation sublayer of layer $\ell$, and sublayer $2\ell$ is its
FFN sublayer. Starting from $\mathcal{H}_0(v)=[h_0(v)]$, the history evolves as
\begin{equation}
  \mathcal{H}_t(v)
  =\mathcal{H}_{t-1}(v)\mathbin{\|}\big[h_t(v)\big],
  \label{eq:history-append}
\end{equation}
where $h_t$ is produced by the AttnRes-augmented sublayer defined below.
The history available immediately before sublayer $t$ is therefore
\begin{equation}
  \mathcal{H}_{t-1}(v)
  =\big[h_0(v),h_1(v),\ldots,h_{t-1}(v)\big].
  \label{eq:history}
\end{equation}
Sublayer $t$ routes $\mathcal{H}_{t-1}$ and appends $h_t$ only after its
transformation. Thus, it can access all earlier contributions but no current
or future output, while the backbone's graph-propagation operator remains
unchanged.

\subsection{Standard Residual Aggregation and Mismatch Hypothesis}

Under standard residual accumulation, the history available before sublayer
$t$ is read as
\begin{equation}
  z_t^{\mathrm{std}}(v)=\sum_{i=0}^{t-1}h_i(v).
  \label{eq:standard-residual-read}
\end{equation}
Although every contribution $h_i(v)$ depends on the node and its graph
context, the aggregation vector is fixed to
$\mathbf{1}\in\mathbb{R}^{t}$ for every node. For comparison of relative
mixture weights, normalizing this vector yields the uniform profile
$t^{-1}\mathbf{1}$. The backbone can adapt the contributions it produces; our
focus is the residual interface after those contributions are fixed. Consider
two nodes $u$ and $v$ that favor distinct normalized history-mixture vectors
$\pi_t(u)\neq\pi_t(v)$ for subsequent computation. No
node-independent vector $w_t$ can equal both. This motivates our
\emph{depth-wise aggregation mismatch} hypothesis: the fixed unit-weight
residual rule can create a depth-wise aggregation mismatch 
when the utility of intermediate computations varies across nodes and contexts.
The vectors $\pi_t$ are conceptual and unobserved; we do
not assume that every node or task requires a distinct mixture.

AttnRes replaces the fixed vector with coefficients
$\boldsymbol{\alpha}_t(v)$ computed from the available history of node $v$.
We use \emph{depth routing} for these coefficients and \emph{node-adaptive
depth aggregation} for the resulting residual interface. We reserve
\emph{depth matching} for the mechanistic interpretation tested in
Section~\ref{sec:depth-matching}, where routing-only interventions examine
whether the learned coefficients are functionally compatible with their
target histories.

\subsection{Node-Adaptive Depth Aggregation with AttnRes}

Following the original AttnRes definition, the operator preceding sublayer
$t$ receives the $t$ candidates in
$\mathcal{H}_{t-1}(v)$.
Each AttnRes operator is parameterized by a learnable query vector $q_t\in\mathbb{R}^d$. 
Within that operator, the same query is used to score every history candidate of every node, 
while node dependence arises from the candidate representations in $\mathcal{H}_{t-1}(v)$. 
Let $\operatorname{RMSNorm}_t$ denote its key normalization~\cite{NEURIPS2019_1e8a1942}. 
For candidate $h_i(v)$, $i\in\{0,\ldots,t-1\}$, the key
$k_{t,i}(v)$ and scalar score $s_{t,i}(v)$ are
\begin{align}
  k_{t,i}(v)
  &=\operatorname{RMSNorm}_t\!\left(h_i(v)\right),
  \label{eq:attnres-key}\\
  s_{t,i}(v)&=q_t^\top k_{t,i}(v).
  \label{eq:attnres-score}
\end{align}
Softmax normalization is applied across the history dimension:
\begin{equation}
  \alpha_{t,i}(v)=
  \frac{\exp\!\left(s_{t,i}(v)\right)}
       {\sum_{j=0}^{t-1}\exp\!\left(s_{t,j}(v)\right)}.
  \label{eq:depth-softmax}
\end{equation}
The original history candidates serve as values. Their weighted sum gives the
routed state
\begin{equation}
  z_t(v)
  =\operatorname{AttnRes}_t\!\left(\mathcal{H}_{t-1}(v)\right)
  =\sum_{i=0}^{t-1}\alpha_{t,i}(v)h_i(v).
  \label{eq:attnres-output}
\end{equation}
The operator has no learned key/value projections, heads, or explicit depth
embeddings. Each AttnRes module adds one node-shared query and one RMSNorm gain vector.
Queries start at zero, giving uniform initial weights. No source-depth encoding
is provided explicitly; any depth preference must arise through candidate
content.

\subsection{Graph Transformer Integration}

Let $N_t$ schematically denote the backbone PreNorm processing~\cite{pmlr-v119-xiong20b} applied between AttnRes and sublayer $t$,
and let $F_t(\cdot;\xi_t)$ denote the original backbone
transformation in sublayer $t$,
where $\xi_t$ denotes any auxiliary backbone context not included in the AttnRes history. 
Suppressing the node argument, the depth interface is
\begin{align}
  z_t
  &=\operatorname{AttnRes}_t\!\left(\mathcal{H}_{t-1}\right),
  \label{eq:sublayer-route}\\
  h_t&=F_t\!\left(N_t(z_t);\xi_t\right).
  \label{eq:sublayer-transform}
\end{align}
For graph-propagation sublayers, 
$\xi_t$ typically contains the graph and structural context required by the original backbone. 
Any auxiliary-state processing or updates follow the original backbone and are omitted from the notation.
Thus, Eqs.~\eqref{eq:sublayer-route}--\eqref{eq:sublayer-transform} modify only
the node-wise depth interface while leaving the backbone's graph-propagation
operator and auxiliary-state update rules unchanged.

In layer $\ell$, $F_{2\ell-1}$ denotes graph propagation and $F_{2\ell}$
the FFN. Their AttnRes placements are Pre-GT and Pre-FFN, respectively. With
$z_\ell^{\mathrm{GT}}$ and $z_\ell^{\mathrm{FFN}}$ denoting the routed inputs,
one complete \method{} layer is
\begin{equation}
\begin{aligned}
  z_\ell^{\mathrm{GT}}
  &=\operatorname{AttnRes}_{2\ell-1}
    \!\left(\mathcal{H}_{2\ell-2}\right),\\
  h_{2\ell-1}
  &=F_{2\ell-1}\!\left(N_{2\ell-1}
    \!\left(z_\ell^{\mathrm{GT}}\right);\xi_{2\ell-1}\right),\\
  \mathcal{H}_{2\ell-1}
  &=\mathcal{H}_{2\ell-2}\mathbin{\|}[h_{2\ell-1}],\\
  z_\ell^{\mathrm{FFN}}
  &=\operatorname{AttnRes}_{2\ell}
    \!\left(\mathcal{H}_{2\ell-1}\right),\\
  h_{2\ell}
  &=F_{2\ell}\!\left(N_{2\ell}
    \!\left(z_\ell^{\mathrm{FFN}}\right);\xi_{2\ell}\right),\\
  \mathcal{H}_{2\ell}
  &=\mathcal{H}_{2\ell-1}\mathbin{\|}[h_{2\ell}].
\end{aligned}
\label{eq:attnres-gt-layer}
\end{equation}
Pre-FFN AttnRes therefore sees the same-layer graph-propagation contribution
$h_{2\ell-1}$, and $\mathcal{H}_{2\ell}$ becomes the next layer's history.

After the $2L$ sublayers, Output AttnRes aggregates the complete causal
contribution history, and its routed state enters the unchanged task head:
\begin{equation}
\begin{aligned}
  z_{\mathrm{out}}
  &=\operatorname{AttnRes}_{\mathrm{out}}(\mathcal{H}_{2L}),\\
  \hat y&=\operatorname{Readout}\!\left(z_{\mathrm{out}}\right).
\end{aligned}
  \label{eq:final-routing}
\end{equation}

\subsection{Block Realization and Complexity}
\label{sec:block-attnres}

Following the Block AttnRes construction in~\cite{chen2026attnres}, we
parameterize the resolution of the available history through a block size
$b\geq1$. It sum-pools $h_1,h_2,\ldots$ in non-overlapping groups
of $b$ consecutive contributions while retaining $h_0$ separately; $b=1$
recovers contribution-level Full AttnRes. Before sublayer $t$, let $m_t$ and
$\rho_t$ denote the number of completed blocks and the size of the active block,
respectively:
\begin{equation}
  m_t=\left\lfloor\frac{t-1}{b}\right\rfloor,
  \qquad
  \rho_t=(t-1)\bmod b.
  \label{eq:block-counts}
\end{equation}
Let $\mathcal B_j=\{(j-1)b+1,\ldots,jb\}$ denote the indices of completed
block $j$, let $B_j$ denote its element-wise sum, and let $P_t$ denote the active
block sum before sublayer $t$:
\begin{align}
  B_j=\sum_{i\in\mathcal B_j}h_i,
  \qquad j\in\{1,\ldots,m_t\}.
  \label{eq:completed-block}
  \\
  P_t=
  \begin{cases}
    \mathbf{0}, & \rho_t=0,\\
    \displaystyle\sum_{i=m_tb+1}^{t-1}h_i, & \rho_t>0,
  \end{cases}
  \label{eq:partial-block}
\end{align}
The resulting candidate history is
\begin{equation}
  \overline{\mathcal{H}}_{t-1}^{(b)}
  =
  \begin{cases}
    [h_0,B_1,\ldots,B_{m_t}], & \rho_t=0,\\
    [h_0,B_1,\ldots,B_{m_t},P_t], & \rho_t>0.
  \end{cases}
  \label{eq:block-history}
\end{equation}
For example, with $b=4$ and nine residual contributions $h_1,\dots,h_9$,
the candidate history is $[h_0,h_1+\cdots+h_4,h_5+\cdots+h_8,h_9]$;
contributions within a pooled candidate do not receive separate depth weights.

After sublayer $t$ produces $h_t$, the block history is updated directly as
\begin{equation}
  \overline{\mathcal{H}}_{t}^{(b)}=
  \begin{cases}
    \overline{\mathcal{H}}_{t-1}^{(b)}
      \mathbin{\|}[h_t], & \rho_t=0,\\
    [h_0,B_1,\ldots,B_{m_t},P_t+h_t],
      & 0<\rho_t<b.
  \end{cases}
  \label{eq:block-update}
\end{equation}
The first case starts an active block; the second updates its running sum and
completes the block when $\rho_t=b-1$.
All candidates use the RMSNorm, query, and softmax in
Eqs.~\eqref{eq:attnres-key}--\eqref{eq:attnres-output}; no block category has
separate scoring parameters. Output AttnRes uses the same candidates.
Because block values are sums, $b$ controls value magnitude, resolution, and
cost; key RMSNorm affects scoring but not values.

Before sublayer $t$, Block AttnRes exposes
$M_t = 1 + \left\lceil \frac{t-1}{b} \right\rceil$
candidates, compared with $t$ under Full AttnRes.
For $n$ nodes and width $d$, the AttnRes cost per call decreases from
$O(ntd)$ to $O(nM_t d)$.
For $t \gg b$, $M_t \approx t/b$, giving approximately a $b$-fold reduction in candidate-wise AttnRes computation.
Across all AttnRes modules, the AttnRes-specific cost decreases from
$O(ndL^2)$
to
$O\!\left(nd\left(\frac{L^2}{b}+L\right)\right)$.
The backbone transformations are unchanged.
Block aggregation introduces no trainable parameters, 
and each AttnRes operator retains one query and one RMSNorm gain vector.

\begin{table*}[!t]
  \centering
  \caption{Test MAE on the \echobenchmark{} (mean $\pm$ standard deviation;
  lower is better). Charge MAE is reported in units of $10^{-3}$.
  Reported rows
  are taken from ECHO; re-tuned rows are our independently tuned backbones.
  Bold and underlined values mark the best and second-best means, respectively.}
  \label{tab:main-results}
  \scriptsize
  \setlength{\tabcolsep}{4pt}
  \resizebox{\textwidth}{!}{\input{tables/tex/main_results_full_echo}}
\end{table*}

%% file: tables/tex/main_results_full_echo.tex
\begin{tabular}{lccccc}
\toprule
Model & \multicolumn{3}{c}{\echosynth{}} & \multicolumn{2}{c}{\echochem{}} \\
\cmidrule(lr){2-4}\cmidrule(lr){5-6}
 & \texttt{diam} $\downarrow$ & \texttt{ecc} $\downarrow$ & \texttt{sssp} $\downarrow$ & \texttt{energy} $\downarrow$ & \texttt{charge} $(10^{-3})$ $\downarrow$ \\
\midrule
A-DGN & $1.151 \pm 0.038$ & $4.981 \pm 0.037$ & $1.176 \pm 0.140$ & $12.486 \pm 1.621$ & $6.543 \pm 0.146$ \\
DRew & $1.243 \pm 0.047$ & $4.651 \pm 0.020$ & $1.279 \pm 0.011$ & $11.325 \pm 2.394$ & $9.086 \pm 0.473$ \\
GCN & $3.832 \pm 0.262$ & $5.233 \pm 0.034$ & $2.102 \pm 0.094$ & $28.112 \pm 1.239$ & $8.421 \pm 0.512$ \\
GCNII & $2.005 \pm 0.093$ & $5.241 \pm 0.030$ & $2.128 \pm 0.429$ & $13.235 \pm 2.630$ & $8.829 \pm 0.021$ \\
GIN & $1.630 \pm 0.161$ & $4.869 \pm 0.092$ & $2.234 \pm 0.271$ & $47.851 \pm 10.154$ & $10.784 \pm 0.059$ \\
GINE & -- & -- & -- & $23.558 \pm 7.568$ & $7.176 \pm 0.371$ \\
GraphCON & $2.969 \pm 0.189$ & $5.474 \pm 0.001$ & $5.734 \pm 0.011$ & $14.295 \pm 0.807$ & $19.629 \pm 0.195$ \\
PH-DGN & $1.627 \pm 0.398$ & $5.068 \pm 0.126$ & $1.323 \pm 0.485$ & $16.080 \pm 1.123$ & $7.915 \pm 0.269$ \\
SWAN & $1.121 \pm 0.070$ & $4.840 \pm 0.045$ & $0.896 \pm 0.232$ & $12.629 \pm 1.157$ & $6.109 \pm 0.103$ \\
\midrule
GRIT (reported) & $\underline{1.014 \pm 0.046}$ & $5.091 \pm 0.158$ & $0.121 \pm 0.013$ & $25.508 \pm 2.507$ & $7.134 \pm 6.090$ \\
GRIT (re-tuned) & $1.323 \pm 0.558$ & $4.819 \pm 0.531$ & $\underline{0.067 \pm 0.001}$ & $20.238 \pm 2.517$ & $6.137 \pm 0.205$ \\
\textbf{\gritattnres{}} & $\mathbf{0.979 \pm 0.051}$ & $4.609 \pm 0.074$ & $\mathbf{0.013 \pm 0.003}$ & $12.029 \pm 0.318$ & $\underline{5.618 \pm 1.219}$ \\
\midrule
GPS (reported) & $2.160 \pm 0.098$ & $4.758 \pm 0.021$ & $0.472 \pm 0.050$ & $\underline{5.257 \pm 0.842}$ & $6.182 \pm 0.219$ \\
GPS (re-tuned) & $1.779 \pm 0.160$ & $\underline{2.493 \pm 1.057}$ & $0.316 \pm 0.090$ & $5.728 \pm 0.674$ & $7.094 \pm 5.996$ \\
\textbf{\gpsattnres{}} & $1.296 \pm 0.168$ & $\mathbf{1.244 \pm 0.033}$ & $0.133 \pm 0.004$ & $\mathbf{3.938 \pm 0.737}$ & $\mathbf{4.665 \pm 0.457}$ \\
\bottomrule
\end{tabular}

%% file: sections/04_experimental_setup.tex
\section{Experiments}

\subsection{ECHO Benchmark}

We use all five \echobenchmark{} tasks~\cite{echobenchmark}. \echosynth{} contains graph-diameter
regression (\texttt{diam}), node-eccentricity regression (\texttt{ecc}), and
single-source shortest-path regression (\texttt{sssp}). \echochem{} contains
density functional theory (DFT)-labeled molecular graphs: graph-level \texttt{energy} predicts the total
energy of each molecule, while node-level \texttt{charge} predicts the partial
charge of each atom. Every task uses mean absolute error (MAE); lower is
better. 
Following ECHO, \texttt{charge} MAE is reported in units of $10^{-3}$.

\subsection{Models, Baselines, and Training Protocol}

We compare \gpsattnres{} and \gritattnres{} with their respective GPS and GRIT
backbones. The AttnRes variants use the recurrence in Section~III while
retaining the original graph-propagation modules. The main table also includes
A-DGN, DRew, GCN, GCNII, GIN, GINE, GraphCON, PH-DGN, and
SWAN~\cite{gravina2023adgn,pmlr-v202-gutteridge23a,kipf2017semi,
pmlr-v119-chen20v,xu2018how,hu2020pretraining,pmlr-v162-rusch22a,
heilig2025porthamiltonian,gravina2025swan}.
For these additional methods, we use the four-seed results reported by
ECHO~\cite{echobenchmark}.

We follow the ECHO tasks, splits, metrics, and validation-MAE model-selection
criterion. To obtain matched backbone comparisons, we independently tune GPS,
GRIT, and their AttnRes variants using the same search domains for shared
backbone hyperparameters and the same trial budgets; the AttnRes search additionally selects the
task-specific block size by validation MAE together with the remaining
hyperparameters. For each task and architecture, we select a
configuration using seed 43 validation MAE, retrain it with seeds
$\{1,2,3,4\}$, and report the mean and standard deviation of test MAE. Each run
uses its best-validation checkpoint; test results do not affect early stopping,
configuration selection, or checkpoint selection. Accuracy ablations retain
the selected main-model configuration, retrain every variant from scratch
with the same four seeds, and do not repeat hyperparameter search.

Experiments were run on NVIDIA A800 GPUs. Selected AttnRes configurations and
code are available in the
\href{https://github.com/yeeFeng233/GraphTransformer_AttnRes}{project repository}.

%% file: sections/05_main_results.tex
\subsection{Main Results}

Table~\ref{tab:main-results} reports the full \echobenchmark{}. Results marked
as reported are taken from ECHO, while the re-tuned rows are GPS and GRIT
independently selected under the same search and evaluation protocol as their
AttnRes variants. Under this matched protocol, \gpsattnres{} and
\gritattnres{} each achieve lower mean MAE than their corresponding re-tuned
backbone on all five tasks.

For a conservative comparison, we use the lower mean MAE of the reported and
re-tuned backbone for each task. Against this stronger
reference, \gpsattnres{} reduces mean MAE by $27.2\%$, $50.1\%$, and $57.9\%$
on \texttt{diam}, \texttt{ecc}, and \texttt{sssp}, and by $25.1\%$ and
$24.5\%$ on \texttt{energy} and \texttt{charge}. The corresponding reductions
for \gritattnres{} are $3.5\%$, $4.4\%$, $80.6\%$, $40.6\%$, and $8.5\%$.
The reductions are therefore consistent in direction, although their magnitude
depends strongly on the backbone and task.

Among all methods in the table, \gritattnres{} obtains the lowest mean on
\texttt{diam} and \texttt{sssp}, while \gpsattnres{} obtains the lowest mean
on \texttt{ecc}, \texttt{energy}, and \texttt{charge}. Thus, at least one
AttnRes variant achieves the lowest mean in every column of
Table~\ref{tab:main-results}. These benchmark results establish the empirical
performance gains; Section~V then examines the proposed depth-aggregation
explanation through controlled interventions and matched ablations.

%% file: sections/06_ablation_analysis.tex
\section{Mechanistic Analysis and Ablations}

We begin with routing interventions on GPS-\texttt{sssp} and
GRIT-\texttt{sssp}, followed by a GPS-\texttt{ecc} replication. Subsequent
ablations examine history-fusion rules, AttnRes placement, performance across
graph-distance ranges, and computational cost using the backbone--task
settings specified in each subsection.

\subsection{Testing Node-Adaptive Depth Matching}
\label{sec:depth-matching}

We intervene on Output AttnRes because it routes the complete causal history
directly to the task head. 
For brevity, let $z(v)\equiv z_{\mathrm{out}}(v)$ and
$\alpha_i(v)\equiv\alpha_{\mathrm{out},i}(v)$ denote the corresponding
Output-AttnRes routing coefficients.
For target node $v$ and donor node $u$ from the same
graph, the original and intervened routed states are
\begin{equation}
  z(v)=\sum_i\alpha_i(v)h_i(v),\qquad
  \widetilde z(v\leftarrow u)=\sum_i\alpha_i(u)h_i(v).
  \label{eq:routing-swap}
\end{equation}
The target candidates $h_i(v)$ remain fixed; only their routing coefficients
are replaced. This directly tests the effect of applying a donor's learned
depth mixture to a fixed target history.

On \texttt{sssp}, nodes are grouped into six target-distance bins. We sweep
the donor bin for every target bin, drawing donors from the same graph. A
cyclic derangement prevents self-donation on the diagonal. To keep the
population identical across donor bins, each matrix uses only nodes eligible
for every donor-bin intervention. Random donor assignment is repeated ten
times within each of four seed checkpoints and averaged within seed. Coverage
of this common population ranges from $57.4\%$ to $99.6\%$ across target bins.
Within each target bin, all compared interventions use the same eligible nodes.
The \texttt{sssp} ground-truth distances supervise the original task; the distance bins are
constructed only for this post-hoc analysis and are neither model inputs nor
routing signals.

\begin{figure*}[t]
\centering
\includegraphics[width=\textwidth]{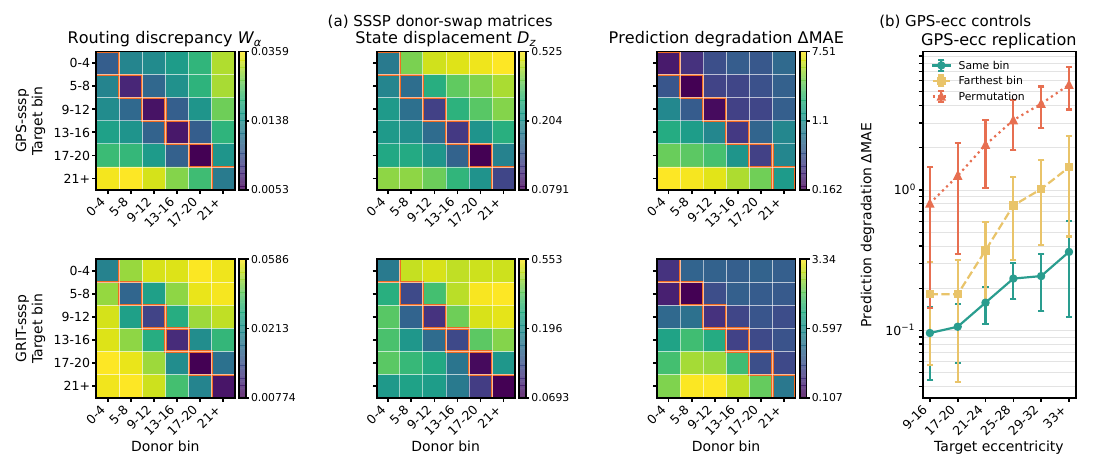}
\caption{Output-routing interventions. (a) SSSP target-bin $\times$
donor-bin matrices for routing discrepancy $W_\alpha$, routed-state
displacement $D_z$, and prediction degradation $\Delta$MAE; orange outlines
mark same-bin swaps. Each panel uses an independent logarithmic color scale.
(b) GPS-\texttt{ecc} $\Delta$MAE for same-bin, farthest-bin, and equal-span
permutation controls. Means are over four seeds; SSSP donor assignment is
additionally averaged over ten repeats. Error bars in (b) show one standard
deviation.}
\label{fig:depth-matching-matrix}
\end{figure*}

We measure three consequences of the swap: the one-dimensional Wasserstein
distance $W_\alpha$ between the original and donor routing distributions over
relative candidate depth, the relative routed-state displacement
$D_z=\lVert\widetilde z-z\rVert_2/
\max(\lVert z\rVert_2,\epsilon)$, and
$\Delta\mathrm{MAE}=\mathrm{MAE}_{\mathrm{intervention}}-
\mathrm{MAE}_{\mathrm{baseline}}$. Here $\epsilon$ is machine epsilon for the
routed-state dtype, and both MAEs use the same eligible target population.
Figure~\ref{fig:depth-matching-matrix} reports their
four-seed means. For $W_\alpha$, $h_0$ is placed at zero; a pooled candidate is
placed at the midpoint of its source-contribution span, normalized by the
total number $2L$ of residual contributions.

On these common eligible populations, same-bin swaps yield the lowest
four-seed mean $W_\alpha$ and $D_z$ in all six target bins for both GPS and
GRIT. They also yield the lowest mean prediction degradation in all six GPS
bins and five of six GRIT bins. Same-bin swaps nevertheless have positive mean
$\Delta$MAE in every target bin, indicating that a shared coarse distance
regime does not make routing fully interchangeable across nodes. Averaging the
six diagonal and 30 off-diagonal matrix cells separately, the
off-diagonal/diagonal ratios for GPS are $2.39\times$ for $W_\alpha$,
$2.38\times$ for $D_z$, and $4.16\times$ for $\Delta\mathrm{MAE}$; the
corresponding GRIT ratios are $3.49\times$, $3.15\times$, and $3.95\times$.
Routing compatibility therefore varies systematically across these
target-defined regimes. For GRIT, the intervention effects are non-monotonic with donor-bin distance,
indicating that graph-distance gap is a contextual grouping variable rather
than a direct scalar measure of routing incompatibility.

We replicate the intervention on GPS-\texttt{ecc}; paired-swap eligibility
ranges from $78.2\%$ to $99.7\%$ across eccentricity bins. Within every bin,
replacing the output weights with those of a node from the farthest
available bin produces a larger mean degradation than a same-bin swap; this
ordering holds in all four seeds. The permutation control keeps $h_0$ fixed
and applies a random nonzero cyclic reassignment within each equal-span group
containing at least two residual candidates; $h_0$ and singleton groups remain
fixed. The intervention preserves both candidate values and the coefficient
multiset. It produces the largest mean degradation in every bin. These
two case studies support the hypothesis that learned routing is functionally
compatible with node-specific histories and associated with target-defined
structural regimes. They do not identify an oracle-optimal mixture or establish
a universal distance-to-depth mapping. These routing-only interventions test
the Output interface directly; the fusion and placement ablations below assess
adaptive aggregation at internal interfaces.

\subsection{History-Fusion Rules}
\label{sec:history-fusion}

We compare five fusion rules on GPS-\texttt{ecc} and GRIT-\texttt{sssp} using
the same block history, placements, and PreNorm operations. Standard residual
uses cumulative summation over all history candidates; latest candidate keeps
only the most recent candidate; and uniform average assigns weight $1/M$ to
each of $M$ candidates. Shared depth weights use module-specific trainable
candidate-position logits shared across nodes, providing a backbone-matched
learned depth-mixing baseline without node conditioning. AttnRes instead computes
node-dependent weights from candidate content. Every variant is trained from
scratch with four seeds. All controls in this subsection and the placement
ablation retain the selected AttnRes configuration without re-tuning;
``Standard residual'' therefore denotes an ablation baseline, distinct from
the independently tuned backbones in Table~\ref{tab:main-results}.

\begin{figure}[t]
\centering
\includegraphics[width=0.92\columnwidth]{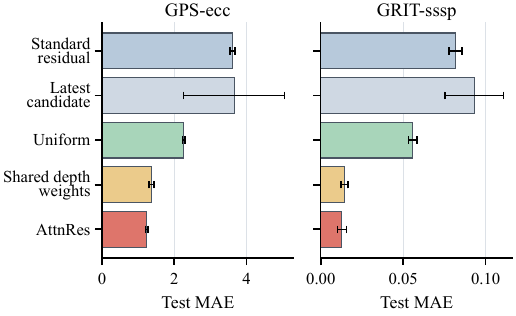}
\caption{History-fusion ablation on GPS-\texttt{ecc} and GRIT-\texttt{sssp}.
All variants use the fixed AttnRes-selected backbone configuration without
re-tuning. Bars show four-seed test MAE (mean $\pm$ standard deviation).}
\label{fig:history-fusion}
\end{figure}

Standard residual retains the complete history through unnormalized cumulative
addition; latest candidate avoids accumulation but discards earlier history.
Uniform averaging retains normalized access to all candidates, while shared
depth weights add a learned but node-independent profile. Among the fixed
fusion rules, uniform averaging has the lowest mean; shared depth weights
reduce the mean further, while content-conditioned AttnRes obtains the lowest
mean in both settings (Fig.~\ref{fig:history-fusion}). Because all five
variants retain the same explicit PreNorm operations, their differences cannot
be attributed to the added PreNorm layers alone. The Output-routing evidence in
Section~\ref{sec:depth-matching} directly tests the functional relevance of
node-specific mixtures at readout. Because shared depth weights
and AttnRes are not nested parameterizations, we do not attribute their gap to
a single design factor.

\subsection{Effect of AttnRes Placement}

We examine Pre-GT, Pre-FFN, and Output AttnRes. Each ablation retains AttnRes
at the other placements and restores standard residual accumulation at the
removed placement. Internal PreNorm normalizes the resulting state before a
backbone transformation, whereas Output AttnRes feeds the task head directly.

\begin{figure}[t]
\centering
\includegraphics[width=\columnwidth]{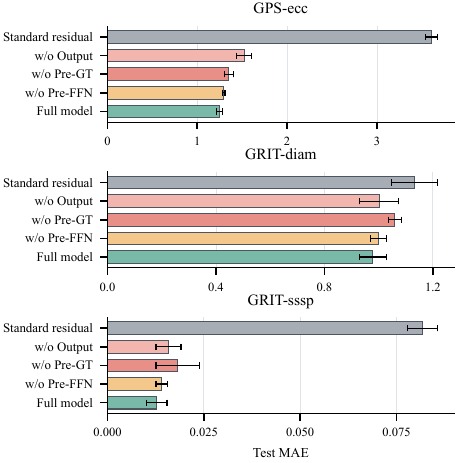}
\caption{Effect of AttnRes placement under the fixed AttnRes-selected
configuration. ``Standard residual'' restores cumulative accumulation at all
three placements; other ablations restore it only at the indicated interface.
Bars show four-seed test MAE; error bars show one standard deviation.}
\label{fig:position-ablation}
\end{figure}

The full model gives the lowest mean in all three settings
(Fig.~\ref{fig:position-ablation}), and removing Pre-GT increases mean MAE in
each. Removing Output AttnRes produces the largest mean increase on
GPS-\texttt{ecc}; on GRIT-\texttt{diam}, the error bars for removing Output
AttnRes and Pre-FFN AttnRes overlap. These controls indicate that adaptive routing can contribute at
multiple stages, although the relative importance of each placement depends
on the backbone and task.

\subsection{Performance across Graph-Distance Ranges}
\label{sec:propagation-range}

\begin{figure}[t]
\centering
\includegraphics[width=\columnwidth]{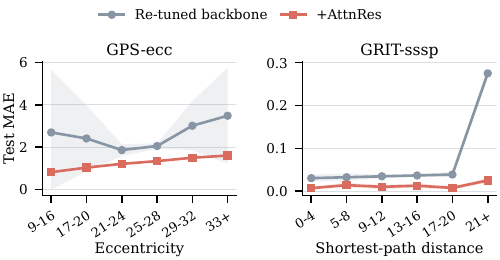}
\caption{Node-level test MAE across graph-distance bins. Lines show
four-seed means and shaded bands one standard deviation. GPS-\texttt{ecc} is
grouped by eccentricity; GRIT-\texttt{sssp} by target shortest-path distance.}
\label{fig:propagation-range}
\end{figure}

We stratify node-level test errors by the graph-distance quantity defining each
target. For \texttt{sssp}, this is ground-truth shortest-path distance. For
\texttt{ecc}, the target is node eccentricity, i.e., the maximum shortest-path
distance from the node to any reachable node. Bins are fixed from target histograms before
examining errors. Figure~\ref{fig:propagation-range} compares four-seed means
for one case study per backbone using the independently re-tuned backbone and
its AttnRes variant.

GPS+AttnRes has a lower four-seed mean MAE in every eccentricity bin, with
reductions of $34.7\%$--$69.7\%$ computed from the displayed four-seed bin
means; the reductions are not monotonic in eccentricity. GRIT+AttnRes likewise
has a lower four-seed mean MAE in every \texttt{sssp} bin, and the $21+$ bin shows
its largest absolute reduction and a $90.8\%$ relative reduction. The gains
therefore extend across the examined distance ranges, while their dependence
on range remains task- and backbone-specific.

\subsection{Computational Cost}

We profile GPS-\texttt{ecc} (19 layers, batch 256) and GRIT-\texttt{sssp}
(48 layers, batch 160) on A800 80 GB GPUs. CUDA-synchronized measurements cover
a complete training epoch and validation pass. Non-OOM configurations average four
process-level repeats rotated across four GPUs; these are system-level repeats, not
accuracy seeds. Backbone rows match the depth, width, and batch size of the
corresponding AttnRes configuration. Full AttnRes on GRIT is evaluated only as
a matched-batch capacity probe.

\input{tables/tex/efficiency_corrected}

Table~\ref{tab:efficiency-corrected} shows that AttnRes is parameter-light but
not activation-light. Block AttnRes adds about $0.3\%$ parameters to GPS and
$0.4\%$ to GRIT. On GPS, $b=2$ increases peak training memory from $27.90$ to
$45.49$~GiB and epoch time from $8.35$ to $12.99$~s; Full AttnRes reaches
$61.05$~GiB and $16.88$~s. Relative to Full AttnRes, block routing reduces
training memory by $25.5\%$ and epoch time by about $23\%$.

The deeper GRIT setting exposes the scaling pressure more clearly. Block
AttnRes with $b=4$ uses $26.18$~GiB and $13.27$~s per epoch, versus $4.73$~GiB
and $5.64$~s for the matched backbone. Full AttnRes exhausts the 80 GB GPU
before completing an epoch, whereas Block AttnRes completes at the same batch
size. These system costs and the task-level accuracy results in
Table~\ref{tab:main-results} are complementary measurements, not a
configuration-matched Pareto comparison; the preferred operating point
depends on memory and runtime budgets.

%% file: tables/tex/efficiency_corrected.tex
\begin{table}[t]
\centering
\caption{Computational cost on an A800 80 GB GPU. Non-OOM configurations average four process-level repeats; GRIT Full AttnRes is a matched-batch feasibility probe. OOM denotes out of memory.}
\label{tab:efficiency-corrected}
\resizebox{\columnwidth}{!}{%
\begin{tabular}{llccccc}
\toprule
Setting & Variant & $b$ & Params (M) & Peak Mem. (GiB) & Epoch (s) & Val. (s) \\
\midrule
GPS-\texttt{ecc} & Matched backbone & -- & 4.545 & 27.90 & 8.35 & 1.06 \\
 & Block AttnRes & 2 & 4.557 & 45.49 & 12.99 & 1.58 \\
 & Full AttnRes & 1 & 4.557 & 61.05 & 16.88 & 2.01 \\
\midrule
GRIT-\texttt{sssp} & Matched backbone & -- & 5.005 & 4.73 & 5.64 & 0.84 \\
 & Block AttnRes & 4 & 5.024 & 26.18 & 13.27 & 1.60 \\
 & Full AttnRes & 1 & 5.024 & OOM & -- & -- \\
\bottomrule
\end{tabular}%
}
\end{table}

%% file: sections/07_conclusion.tex
\section{Conclusion}

We analyze the fixed unit-weight aggregation imposed by standard residual accumulation in Graph Transformers
and hypothesize that this rule can create a depth-wise aggregation mismatch when the utility of intermediate computations varies across nodes and contexts.
To study this problem, we formalize graph-propagation and FFN contributions 
as a causal node-wise history and instantiate node-adaptive depth aggregation with AttnRes 
while preserving the backbone's graph-propagation machinery. 
On all five ECHO tasks, both \gpsattnres{} and \gritattnres{} achieve lower mean MAE 
than their independently tuned backbones, 
with at least one variant obtaining the lowest mean among the compared methods on every task.

Controlled routing interventions and matched ablations further support the view 
that learned depth mixtures are node-specific, functionally relevant, 
and associated with structural context. 
These findings support node-adaptive depth aggregation as a useful complement to cross-node communication 
for long-range graph reasoning. 
Our empirical scope is limited to GPS and GRIT on ECHO, and AttnRes incurs
additional activation memory and runtime; broader evaluation across architectures and benchmarks remains future work.